\documentclass[letterpaper, 10 pt, conference]{ieeeconf}  

\IEEEoverridecommandlockouts                              

\usepackage{caption}
\usepackage[ruled]{algorithm2e}
\newcommand{\nextnr}{\stepcounter{AlgoLine}\ShowLn}
\usepackage{algpseudocode}
\algnewcommand\algorithmicforeach{\textbf{for each}}
\algdef{S}[FOR]{ForEach}[1]{\algorithmicforeach\ #1\ \algorithmicdo}
\SetKwRepeat{Do}{do}{while}

\usepackage{booktabs} 
\usepackage{graphicx} 
\usepackage{makecell} 
\usepackage{amsmath} 
\usepackage{amssymb} 
\usepackage{xcolor,blindtext} 
\usepackage{subfig}
\usepackage{color} 
\usepackage{balance} 
\usepackage{multirow}

\begin{document}

\title{VascularPilot3D: Toward a 3D fully autonomous navigation for endovascular robotics}

\author{Jingwei Song, Keke Yang, Han Chen, Jiayi Liu, Yinan Gu, Qianxin Hui, Yanqi Huang, Meng Li, Zheng Zhang, \\
Tuoyu Cao, and Maani Ghaffari%
\thanks{J. Song is with United Imaging Research Institute of Intelligent Imaging, Beijing 100144, China
 \texttt{jingweisong.eng@outlook.com}}%
 \thanks{K. Yang, H. Chen, J. Liu, Y. Gu, Q. Hui, Y. Huang, M. Li, Z. Zhang, and T. Cao are with United Imaging, Shanghai, China. \texttt{\{keke.yang, han.chen03, jiayi.liu01, yinan.gu, qianxin.hui, yanqi.huang, meng.li02, zheng.zhang, tuoyu.cao\}@united-imaging.com}. }
   \thanks{M. Ghaffari is with the University of Michigan, Ann Arbor, MI 48109, USA. \texttt{maanigj@umich.edu}.}%
}
\maketitle
\thispagestyle{empty}
\pagestyle{empty}

\begin{abstract}

This research reports VascularPilot3D, the first 3D fully autonomous endovascular robot navigation system. As an exploration toward autonomous guidewire navigation, VascularPilot3D is developed as a complete navigation system based on intra-operative imaging systems (fluoroscopic X-ray in this study) and typical endovascular robots. VascularPilot3D adopts previously researched fast 3D-2D vessel registration algorithms and guidewire segmentation methods as its perception modules. We additionally propose three modules: a topology-constrained 2D-3D instrument end-point lifting method, a tree-based fast path planning algorithm, and a prior-free endovascular navigation strategy. VascularPilot3D is compatible with most mainstream endovascular robots. Ex-vivo experiments validate that VascularPilot3D achieves $100\%$ success rate among 25 trials. It reduces the human surgeon's overall control loops by $18.38\%$. VascularPilot3D is promising for general clinical autonomous endovascular navigations.
\end{abstract}


\IEEEpeerreviewmaketitle

\section{Introduction}
    Robot-assisted Endovascular Image-Guided Interventions (EIGIs) are gaining popularity in contrast to conventional regular EIGIs in treating heart disease, cancer, stroke, or neurovascular disease due to their lower mortality rate, faster recovery, and most importantly, limited exposure to radiation for surgeons~\cite{da2020challenges}. Existing robot-assisted EIGIs follow the master-slave robot operation system and are manipulated manually by surgeons for guidewire/catheter navigation and other operation procedures~\cite{pore2023autonomous}. Pre-operative Computed Tomography (CT) or Computed Tomography Angiography (CTA) and intra-operative fluoroscopic X-ray or Digitally Subtracted Angiograms (DSA) serve as maps and eyes for surgeons' navigation tasks. In clinical applications, hand-operated navigation is laborious for surgeons because humans have difficulty adapting to multi-modal data, soft-tissues' deformation, non-isometric friction, and incompatibility of control panels~\cite{simaan2018medical}. \textbf{Therefore, this work proposes the first 3D fully autonomous endovascular navigation system VascularPilot3D, under intra-operative (fluoroscopic X-ray in this study) and pre-operative data (CTA in this study).} VascularPilot3D aims to reduce time consumption and the mental and physical burden placed on the operators in navigation procedures, allowing more focus on the clinical aspects. \par 

    Existing autonomous endovascular robots navigated instruments through intelligent agents which were trained with Reinforcement Learning (RL)~\cite{fagogenis2019autonomous,behr2019deep,zhao2019cnn,chi2020collaborative,karstensen2020autonomous,kweon2021deep,karstensen2022learning,schegg2022automated}. Among these RL-based studies,~\cite{behr2019deep,zhao2019cnn,chi2020collaborative,kweon2021deep} adopted imitation learning to make full use of human experiences while others were fully unsupervised. These works were trained and tested on phantoms except~\cite{fagogenis2019autonomous} whose ``haptic vision'' setup falls out of this article's scope. It should be pointed out that clinical applications of these RL-based approaches require a massive amount of training on phantoms or patients, and generalization capability across different patients is unknown. More importantly, all these works were conducted in 2D space. According to our knowledge, no researches discuss the vessels‘ topological ambiguity issues in 2D navigation tasks. It is obvious that some trafficable intersections on 2D images were illusions from non-intersect overlapping 3D vessels. Thus, a 3D navigation method may overcome topological ambiguity and avoid expensive or even inaccessible RL-based navigation training. Meanwhile, we also aim at a a prior-free system that requires no heavy amount of navigation training like RL studies.\par

This research proposes VascularPilot3D as the first 3D prior-free autonomous endovascular navigation system. It follows the typical autonomous driving's ``mapping-localization-navigation'' workflow, where ``mapping'' builds the full 3D vascular from pre-operative data, ``localization'' localizes instruments intra-operatively, ``navigation'' plans the global route, and makes local sequential controls. Among them, ``mapping'' and ``localization'' methods have been widely analyzed. Massive amount of studies have shown that 3D U-Net like Deep Neural Networks (DNNs)~\cite{cciccek20163d,yang2019discriminative,chen2019coronary,lei2020automated,pan2021coronary,wu2023transformer,zeng2023imagecas,mu2023attention} and graph convolution network~\cite{kong2020learning,zhao2022graph,li2022dual} achieve satisfying and robust 3D vessel ``mapping''. Meanwhile, we propose that ``localization'' can be realized by combining fast 2D instrument segmentation and 3D-2D vessel registration. 2D instrument segmentation, similar to 3D vessel segmentation, is developed taking the off-the-shelf lightweight U-Net DNN. Prior-free fast 3D-2D vessel registration algorithms~\cite{aylward2003registration,markelj2008robust,steininger2012auto,rivest2012nonrigid,song2024iterative} have also been well studied. \par 
  
\begin{figure*}[!h]
		\centering
		\subfloat{
			\begin{minipage}[]{1\textwidth}
				\centering
				\includegraphics[width=1\linewidth]{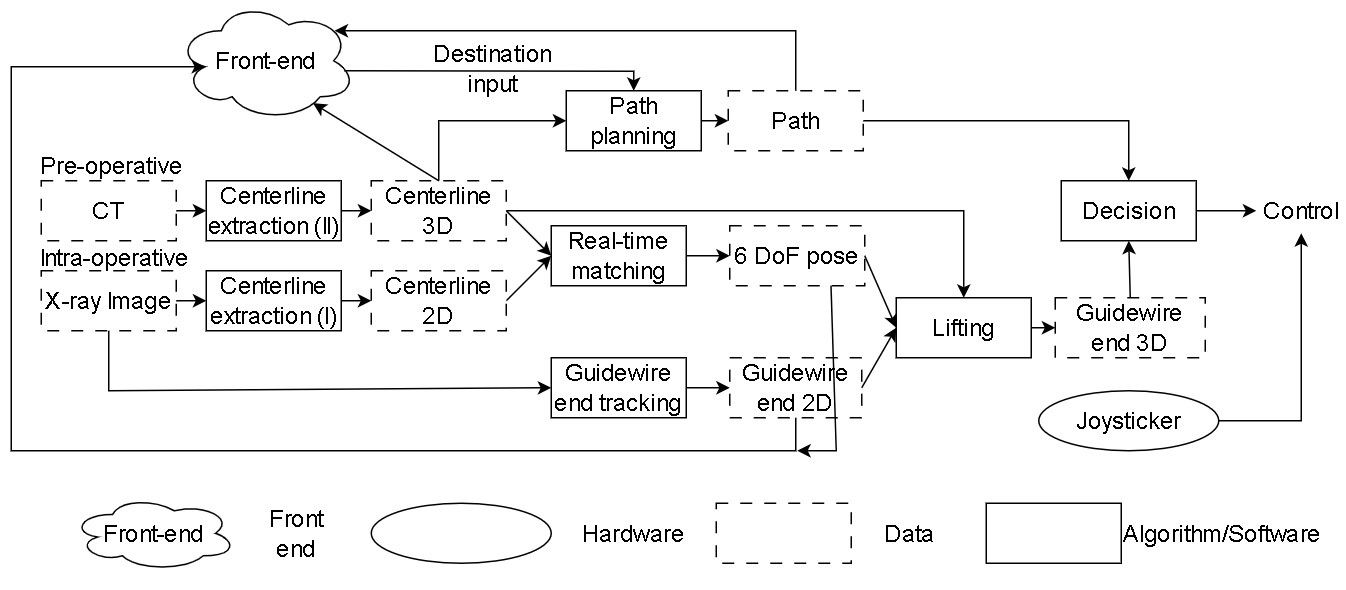}
			\end{minipage}
		}
		\caption{Presented is the framework of the proposed VascularPilot3D. VascularPilot3D involves centerline extraction, 3D-2D matching, guidewire endpoint tracking, path planning, 2D-3D lifting, and decision modules. The framework is a general loop control process. After initialization, VascularPilot3D uses intra-operative data and feedback to control the instrument.}
		\label{fig_framework}
	\end{figure*}  
 
Our VascularPilot3D leverages previous algorithms for ``mapping'' and ``localization'' and enhances the localization with 3D topology constraints. More importantly, a newly proposed prior-free ``navigation'' module achieves efficient global path planning by presenting the vessel as a tree data structure with parent pointers. Additionally, we summarize the path and instrument states as 4 scenarios and propose a practical ``backward-and-forward'' control strategy that covers all scenarios. Our contributions include:

\begin{itemize}
\item To our knowledge, VascularPilot3D is the first 3D prior-free autonomous endovascular navigation system. 3D navigation overcomes the false intersection problem that occurs in conventional 2D navigation works.
\item A novel topology-constrained instrument endpoint 2D to 3D registration method is proposed. The 3D points can be accurately tracked based on the tracked 2D point.
\item A tree-based map presentation and global path planning method is proposed. A practical local control strategy has also been developed for fast navigation.
\item Ex-vivo experiments demonstrate that VascularPilot3D consumes $18.38\%$ less control loops than human operators and achieves $100\%$ success rate among 25 trials.
\end{itemize}

 \begin{figure}[t]
    \centering
    \includegraphics[width=1\columnwidth]{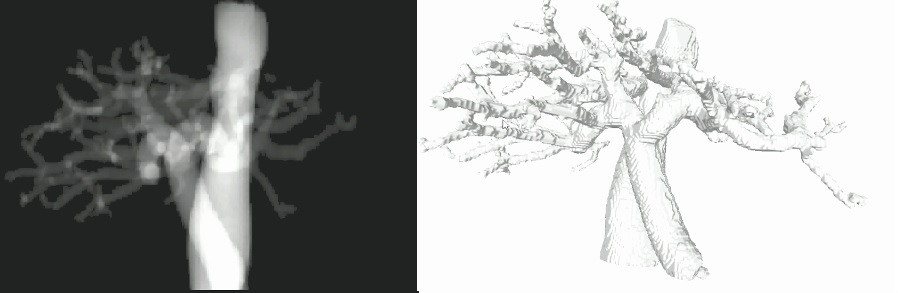}
    \caption{Left is the sample 2D DSA image projected from its 3D vessel, which is shown on the right. It shows that the projected vessels bring heavy topological ambiguities in navigation tasks.}
    \label{fig_vessel_overlap}
\end{figure}
	\section{Methodology}
	
	\subsection{System overview}
	\label{section_2_1}
	
Fig. \ref{fig_framework} shows the general framework of VascularPilot3D. ``Centerline extraction (I)'' yields intra-operative 2D vessel centerlines in real-time from X-ray images. ``Centerline extraction (II)'' extracts 3D vessel centerlines from the pre-operative data before the operation. Then, a real-time 3D-2D registration is performed to align 3D and 2D vessels. Meanwhile, a lightweight DNN is applied to estimate the endpoint position of the instrument (guidewire in this study). With the correct 3D-2D registration, the 2D endpoint is lifted from 2D to 3D space with the correct vessel topology. Finally, a global path is planned from the current 3D endpoint position to the target 3D point chosen manually by the operator in the path planning phase. VascularPilot3D then makes iterative control based on the estimated endpoint's 3D position from the latest obtained intra-operative 2D image. In all, VascularPilot3D takes pre-operative and intra-operative images as feedback and controls the instruments.

\subsection{Centerline segmentation and 3D-2D registration}
\label{sec_irls_cvo}

This work strictly follows~\cite{song2024iterative} for 3D and 2D centerline segmentation as well as non-rigid registration, and we briefly introduce the procedure for completeness. Widely used Otsu's method~\cite{otsu1979threshold} was applied for fast 2D vessel segmentation. A DNN-based method from commercial software was used for an offline 3D vessel segmentation. It should be emphasized that other off-the-shelf prior-based DNNs~\cite{li2022dual,moccia2018blood} are also applicable for 3D/2D image segmentation. Based on the segmented 3D and 2D centerlines, registration (optimal rigid and nonrigid transformation) was achieved by jointly maximizing Reproducing Kernel Hilbert Space (RKHS) sum constrained by regularization. After registration, point-wise 3D-2D vessel associations can be retrieved. The RKHS-based optimization function is

    \begin{equation}
\begin{aligned}
	\label{GMM_object_func}
	&
 \operatorname{max}_ {\mathbf{T},\theta}  E_{data} , \text{ such that},
 \operatorname{min}_ {\mathbf{T}, \theta} \lambda_1 E_{init} + \lambda_2 E_{reg}\\
 &E_{data}= \sum_{i \in \Omega}\mathrm{w}_i\sum_{j \in \Delta(i)}\\
 &\exp \left(-\frac{\lVert \pi(\mathbf{T}\Gamma(\mathbf{T}_{dsa\_cb}\mathbf{p}_i-\mathbf{c},\theta),\mathbf{K})-\mathbf{q}_j \rVert^2_2}{2\ell^2}\right)\\
  &E_{init}= \lVert \operatorname{log}(\mathbf{T}^{-1}_0\mathbf{T})^{\vee} \rVert_2^{2}\\
  &E_{reg} = \mathrm{w}_1 \sum_{i \in \Omega} \left\lVert \mathbf{r}_i \right\lVert^2_{2}  +  \mathrm{w}_2 \sum_{i \in \Omega}\sum_{j\in \{i-1,i+1\}} \left\lVert \mathbf{r}_i - \mathbf{r}_j \right\lVert^2_{2} + \\
  &\mathrm{w}_3 \sum_{i \in \Omega}\sum_{j \in \Omega_i} \left\lVert \mathbf{r}_i - \mathbf{r}_j \right\lVert^2_{2},
\end{aligned}
\end{equation}

\noindent where $\mathbf{T}_{dsa\_cb} \in \mathrm{SE}(3)$ is the pre-calibrated transformation matrix from pre-operative coordinate (CT for example) to intra-operative coordinate (DSA for example), $\Omega$ is the set of 3D centerline point $\mathbf{p}_i \in \mathbb{R}^4$ (in homogeneous coordinate and the same applies to the rest), $\Delta(i)$ is the set of corresponding 2D point $\mathbf{q}_j \in \mathbb{R}^3$, $\mathbf{T}_0$ is the initial pose or pose in previous state and $\ell$, $\lambda_1$, $\lambda_2$, $\mathrm{w}_1$, $\mathrm{w}_2$, $\mathrm{w}_3$ are hyperparameters, and $\theta = \{\mathbf{r}_i\}, \mathbf{r}_i \in \mathbb{R}^4$ (last dimension is fiexed to $0$) is the deformation parameter, $\mathbf{K} \in \mathbb{R}^{3\times4}$ is the camera intrinsic matrix following pin-hole camera. $E_{init}$ defines the distance on $\mathrm{SE}(3)$ manifold and $(\cdot)^\vee$ converts Lie algebra $\mathfrak{se}(3)$ to 6-vector $\mathbb{R}^6$. $E_{init}$ is used to avoid scale drift. In the first step of minimization, $\ell$ is initialized as the maximum $||\pi(\mathbf{T}\Gamma(\mathbf{T}_{dsa\_cb}\mathbf{p}_i-\mathbf{c},\theta))-\mathbf{q}_j,\mathbf{K}||^2_2$ and shrinked by half every 5 iterations. This work follows the latest progress~\cite{clark2021nonparametric} and adopts the Iterative Reweighted Least Squares (IRLS) combined with Levenberg–Marquardt for fast optimization.\par

\begin{algorithm}[t]
	\caption{Strategy-based navigation.}
	\label{Algorithm_nav}
	\KwIn{3D start $\mathbf{p}_s \in \mathbb{R}^3$, 3D target $\mathbf{p}_e \in \mathbb{R}^3$ and $\mathcal{M}(\cdot)$ tests if input is on path} 
	\KwOut{Sequential translation and rotation control $\mathbf{C}=\{\cdot,\cdot\}$, both elements are in $\mathbb{R}^1$}
    \nextnr
	Set \textit{Flag\_back} and \textit{Flag\_on\_Path\_Last} as \textit{True} \& $\mathrm{w} = 0$\\
    \nextnr
    \While{$\mathcal{D}(\mathcal{L}(),\mathbf{p}_e) >\mathrm{r}_{th}$}{
        \nextnr
		\eIf{\textit{Flag\_back} is \textit{True}}{
        \nextnr
            \eIf{$\mathcal{L}()$ is \textit{True}}{
            \nextnr
                $\mathbf{C}=\{-10,0\}$/* Backward */ 
            }
            {
                \nextnr
                $\mathbf{C}=\{10,0\}$/* Forward */ \\
                \nextnr
                Set \textit{Flag\_back} as \textit{False} \& 
                Replan path from $\mathcal{L}()$ to $\mathbf{p}_e$ 
            }
        }
        {
            \nextnr
            \If{$\mathrm{w} > \mathrm{w}_{th}$}{
                \nextnr
                Replan path from $\mathcal{L}()$ to $\mathbf{p}_e$ \&
                $\mathrm{w} = 0$ \& \\
                \nextnr
                Set \textit{Flag\_back} as \textit{True} /* Redo path planning */
            }
            \nextnr
            \eIf{$\mathcal{L}()$ is \textit{True}}{
                \nextnr
                $\mathrm{w} = 0$\\
                \nextnr
                \eIf{\textit{Flag\_on\_Path\_Last} is \textit{True}}{
                    \nextnr
                    $\mathbf{C}=\{\mathrm{c},0\}$/* Forward */
                }{
                    \nextnr
                    $\mathbf{C}=\{\mathrm{c},1\}$/* Forward and rotate */\\
                }
            }
            {
                \nextnr
                $\mathbf{C}=\{-\mathrm{c},1\}$ \& $\mathrm{w} = \mathrm{w} + 1$ /* Backward and rotate */
            }
        }
	}
	Notation: $\mathcal{D}(\cdot,\cdot)$ measures the Euclidean distance between two points. $\mathrm{r}_{th}$ determines whether the target has been reached. $\mathrm{w}$ records step numbers. $\mathrm{w}_{th}$ controls if the path should be replanned. $\mathbf{C}$ is sent to slave machines each time it is assigned. $\mathrm{c}$ is a random number ranging from 8 to 12. $\mathcal{L}()$ localizes the endpoint 3D position following Section \ref{section_2D_3D_mapping}.\\
\end{algorithm}

\begin{figure}[t]
    \centering
    \includegraphics[width=1\columnwidth]{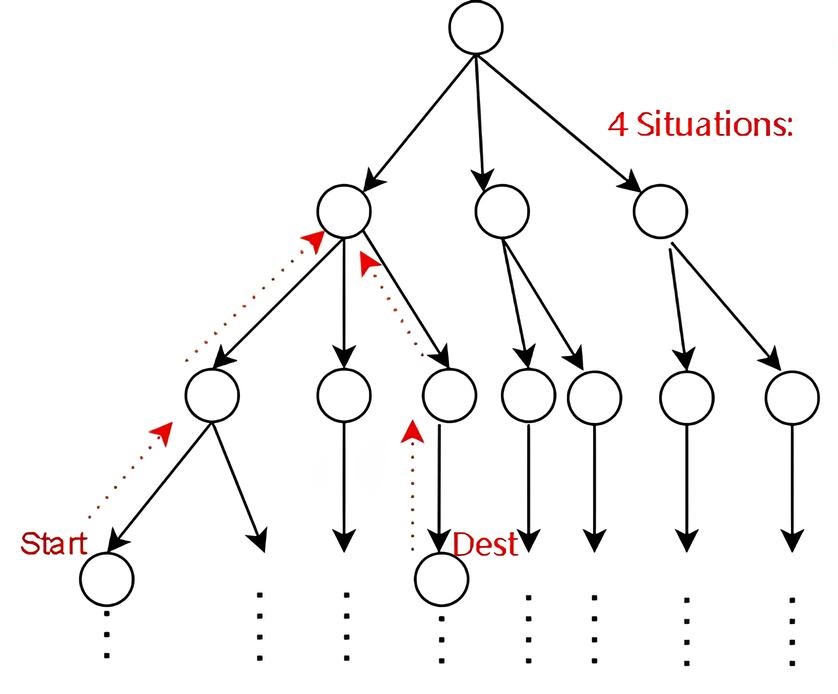}
    \caption{This figure shows the efficient global path planning by tracing both nodes to their common parent. ``Dest'' stands for the destination. The nodes, or circles, are the intersections of vessels.}
    \label{fig_nav_vis_global}
\end{figure}

\begin{figure}[t]
    \centering
    \includegraphics[width=1\columnwidth]{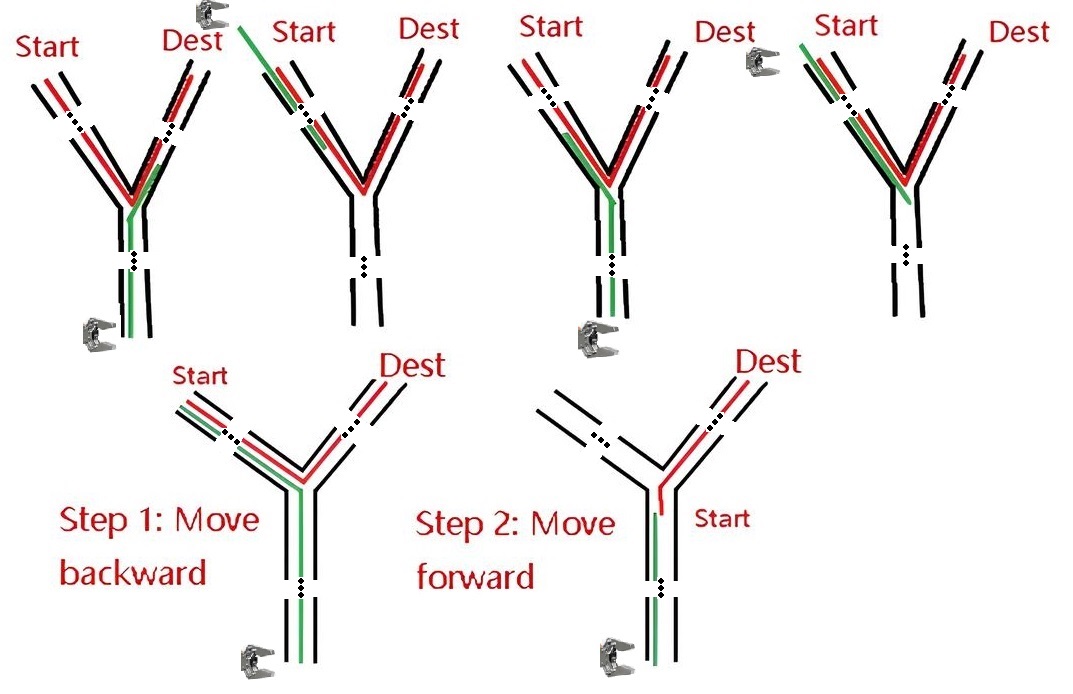}
    \caption{This figure visualizes the idea of local path planning''. ``Dest'' stands for the destination. The upper figure reveals the general four scenarios including the setup of path, instrument, and manipulator. Although the vessel contains only one bifurcation, all other bifurcations can be embedded and simplified as dots within each branch. The bottom shows the reason for our ``backward-and-forward'' manner. Red curves are the paths from ``Start'' to ``Dest''. Green curves are the instruments. }
    \label{fig_nav_vis_local}
\end{figure}

\subsection{Instrument endpoint estimation and 2D to 3D endpoint lifting}
\label{section_2D_3D_mapping}
During the operation, real-time continuous X-ray images are obtained, and the instrument's 2D endpoint can be estimated with UNet-like DNN methods. SA-UNet~\cite{guo2021sa}, first built for segmenting retinal vessels, is adopted for instrument segmentation as it achieves state-of-the-art performance on limited training data in curve-like segmentation scenario. \cite{zhang1984fast}'s skeletonization method is applied to instrument mask extraction due to its efficiency and accuracy. Then, the skeleton points that have one adjacent are presumed to be candidate endpoints. The surgeons must manually choose the endpoint if more than one candidate is found in the first frame; the selected endpoint is then adopted as the seed for the iterative tracking process.\par

With the real-time 3D-2D alignment discussed in Section \ref{sec_irls_cvo} and 2D endpoint tracking, the estimated 2D endpoint is then lifted to 3D in the pre-operative vessel's coordinate. Although computer vision community agrees that lifting 2D to 3D is ambiguous in viewing direction, we overcome the issue using vessel registration constraints. Since the 3D endpoint and its 2D projection are all within the vessel, the 2D endpoint is lifted to 3D space via its closest 2D vessel point and the 2D vessel point's corresponding 3D vessel point. It is obvious that the accuracy of 3D localization depends on the registration and vessel radius.\par  
 
\subsection{Global and local path planning}


Previous 2D navigation methods' global and local path planning modules~\cite{fagogenis2019autonomous,behr2019deep,zhao2019cnn,chi2020collaborative,karstensen2020autonomous,kweon2021deep,karstensen2022learning,schegg2022automated} cannot identify and handle false 2D intersections, as illustrated in Fig. \ref{fig_vessel_overlap}. Moreover, \cite{karstensen2022learning,kweon2021deep,karstensen2020autonomous,chi2020collaborative,zhao2019cnn} pointed out that the state presented in absolute 2D position makes the trained network difficult to be generalized. We agree with~\cite{karstensen2022learning} that the agent ``learns the movements to navigate this specific geometry''. Therefore, this research unifies global and local path planning on 3D space to circumvent false intersections and enable more generalized navigation.\par  

\textbf{Global path planning}, or the route from the start position to the target position, is done on the 3D map in the multiway trees data structure. To fasten the search procedure, each node in the multiway trees has an additional pointer to its parent. During global path planning, the 3D endpoint position (in Section \ref{section_2D_3D_mapping}) is the start point, and the surgeon chooses the 3D target position. The route is then efficiently built by tracing both nodes to their common parent in the multiway trees (shown in Fig. \ref{fig_nav_vis_global}). This achieves the time complexity as $\mathcal{O}((\log{}n)^2)$ comparing to Dijikstra's $\mathcal{O}(N^2)$. It should be noted that our approach only handles loop-free route searching compared to Dijikstra.\par

\textbf{Local path planning}, or control, refers to sending instrument translation and rotation motion control to the slave machines. Although existing RL-based methods achieve successful navigation through iterative RL training, applying RL to 3D navigation should solve the following obstacles considering hardware, cost, and ethics: fast convergence in the phantom training process, fast adaptation across modals, efficient inference, and uncertainties. Uncertainties, mainly caused by unreliable localization, hardware control in nonrigid and nonisometric friction, and system delay, bring unpredictable phenomena in obtaining the correct control. Our experiments reveal that RL behaves well in a virtual environment and deteriorates significantly in real-world phantom. Thus, we gave up the previous RL roadmap and sought a prior-free strategy-based method. Inspired by surgeon's ``trial-and-error'' in their routine endovascular navigation tasks~\cite{duran2015kinematics,mazomenos2016catheter}, VascularPilot3D summarized the process and transplanted the experience as a simple yet effective strategy as shown in Algorithm~\ref{Algorithm_nav}. \par

Fig. \ref{fig_nav_vis_local}  reveals the idea behind Algorithm~\ref{Algorithm_nav}. Although Algorithm~\ref{Algorithm_nav} is complete and re-implementable, Fig. \ref{fig_nav_vis_local} compliments with a visualization of the idea behind Algorithm~\ref{Algorithm_nav}. All navigation setups between path and instrument can be summarized as four scenarios shown in the upper of Fig. \ref{fig_nav_vis_local}. The four scenarios can be summarized in a ``backward-and-forward'' manner (bottom of Fig. \ref{fig_nav_vis_local}). As shown in Algorithm~\ref{Algorithm_nav}, our navigation strategy first moves the instrument backward until the endpoint falls out of the planned path. Then, it redo path planning by moving the instrument forward to the destination. \textbf{All navigation can be handled in ``backward-and-forward'' manner.} Readers can validate the claim following Fig. \ref{fig_nav_vis_local}.\par 

\begin{figure}[t]
    \centering
\includegraphics[width=1\columnwidth]{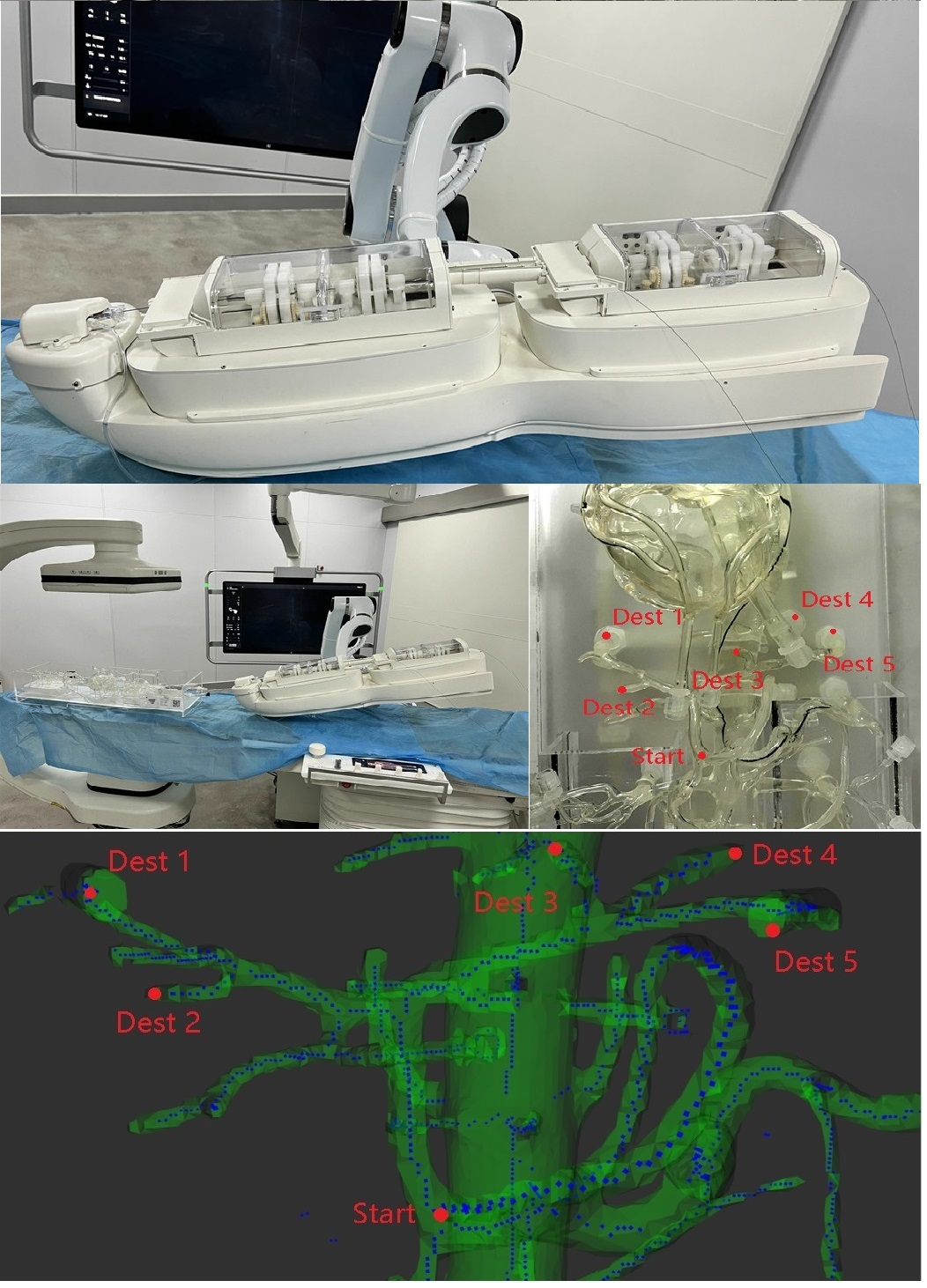}
    \caption{The figure shows the experiment setup, robot, phantom, and 5 tasks. ``Dest'' stands for the destination. }
    \label{fig_5_tasks}
\end{figure}

    \begin{figure*}[!h]
		\centering
		\subfloat{
			\begin{minipage}[]{1\textwidth}
				\centering
				\includegraphics[width=1\linewidth]{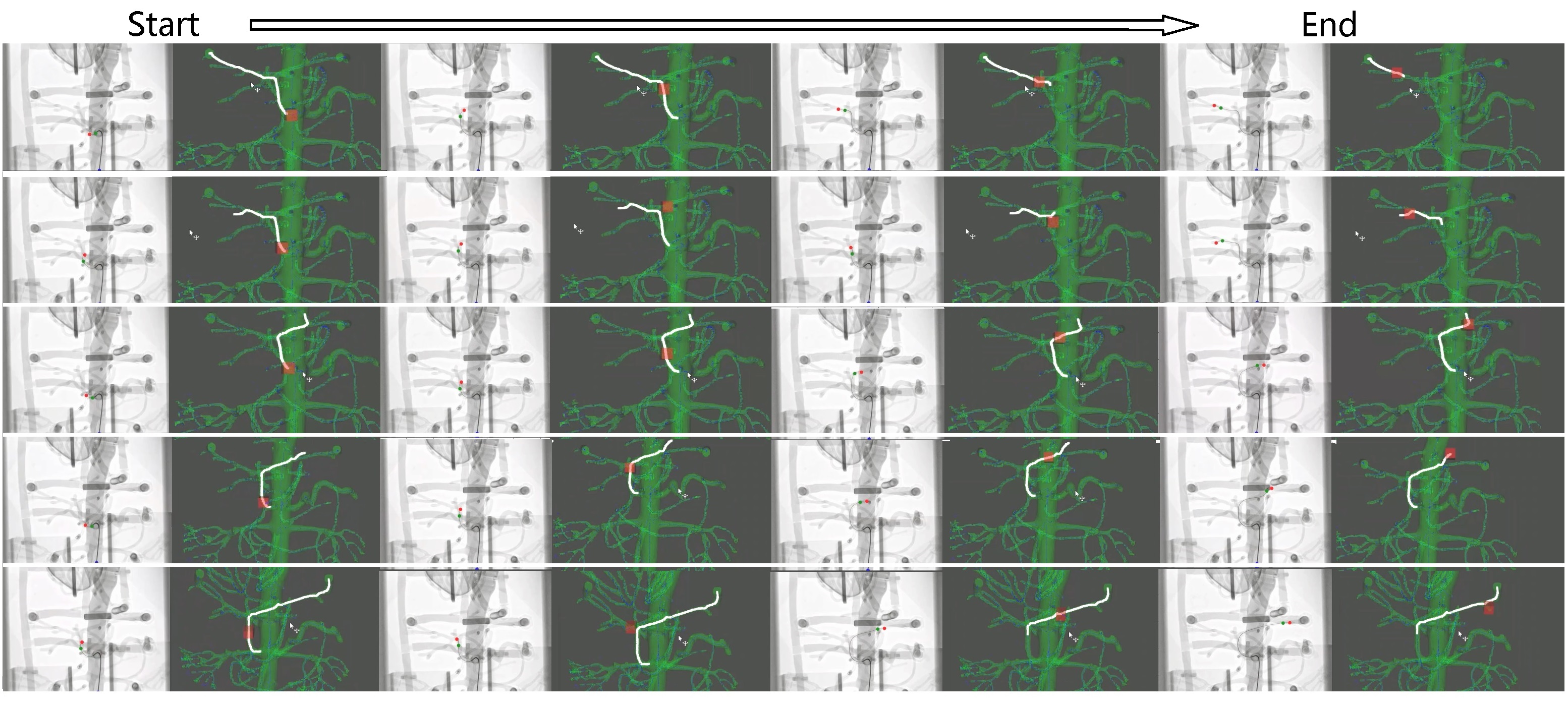}
			\end{minipage}
		}
		\caption{Figure shows the sample navigation of VascularPilot3D on the five tasks (each row shows one task). The red marker is the guidewire endpoint on both 2D and 3D. The white curve is the planned 3D path. Both 2D and 3D spaces are synchronized. \textbf{Full navigations can be found in the attached video.}}
		\label{fig_exvivo_dataset}
	\end{figure*}

	\section{Results and discussion}

    VascularPiolot3D was implemented as image receiver, 3D-2D registration, 2D endpoint extraction, and navigation packages and mounted on Robot Operating System~\cite{quigley2009ros}. The complete software was deployed on a commercial laptop ALIENWARE M17 R4 (i7-10870H, 32Gb RAM, and GeForce RTX 3060 6Gb). 3D-2D registration package was implemented in C++ for efficient registration, and the rest were implemented in Python. Our own robot was built, which carries VascularPiolot3D (see Fig. \ref{fig_5_tasks}). The proposed VascularPiolot3D is modular to other widely used vascular robots as well. Pre-operative CT data of a 3D phantom was collected using United Imaging's uCT 860 with volume size $512\times512\times1959$ and voxel size $0.824\times0.824\times0.5 mm$. VascularPiolot3D grabbed real-time X-ray images from United Imaging's uAngio 960. All collected 2D images were resized as $512 \times 512$. The average pixel size is in $0.30 mm$. The average delay from uAngio 960 to VascularPiolot3D is $0.2 s$. The hyperparameters $\lambda_1$, $\lambda_2$, ${w}_1$, ${w}_2$ and ${w}_3$ are set as $100$, $1$, $0.1$, $10$ and $1$. \par

    $527$ fluoroscopic X-ray images were collected and labeled to train the instrument segmentation network SA-UNet. All these images were captured with uAngio 960 in the same mode, and instruments were manually labeled. Among them, $450$ pairs were used for training, and $77$ pairs for validation. After the training, the network is applied to extract the instrument endpoints.\par  
    
    VascularPilot3D was tested on a phantom that performs navigation on 5 different tasks shown in Fig. \ref{fig_5_tasks}. 5 trials were conducted in each task, and the average values and standard deviations were recorded. Since chosen tasks are typical tasks and VascularPiolot3D is prior-free, successful navigation can validate that VascularPiolot3D is suitable for any task without adaptation and training. It should also be noted that the major claim is the performance of full functionality of VascularPilot3D, and thus, validation is focused on navigation performance. Validation of 3D-2D registration can be found in~\cite{song2024iterative}.

 \begin{table*}[]
		\centering
		\caption{This table compares the number of control loops of VascularPilot3D, human operator and a typical RL~\cite{behr2019deep} on the 5 tasks. ``Mean'' and ``std'' are the average and standard deviation of control loops. ``-'' indicates failing in all trials and ``*'' fails some trials.}
  
		\begin{tabular}{lrrrrrrrrrr}
\toprule
\multirow{2}{*}{Method} & \multicolumn{2}{c}{Task1}        & \multicolumn{2}{c}{Task2}        & \multicolumn{2}{c}{Task3}        & \multicolumn{2}{c}{Task4}        & \multicolumn{2}{c}{Task5}        \\ \cline{2-11} 
                        & \multicolumn{1}{l}{Mean} & Std   & \multicolumn{1}{l}{Mean} & Std   & \multicolumn{1}{l}{Mean} & Std   & \multicolumn{1}{l}{Mean} & Std   & \multicolumn{1}{l}{Mean} & Std   \\ \midrule
VascularPilot3D         & 31.14                    & 25.98 & \textbf{24.50}                    & 10.91 & \textbf{21.90}                    & 6.75  & 68.50                    & 43.15 & 30.65                    & 6.70  \\
Surgeon                 & \textbf{29.89}                    & 11.70  & 51.46                    & 20.93 & 31.65                    & 11.46 & 66.25                    & 23.00 & 37.23                    & 11.56 \\
RL navigation~\cite{behr2019deep}                 & 53.67                    & 31.04  & -       & -       & 40.67                    & 21.04 & \textbf{57.00}                    & * & 
\textbf{26.33}                    & 8.11 \\\bottomrule
\end{tabular}
		\label{Table_exvivo_dataset}
	\end{table*}
    
    \subsection{Navigation performances}

    Fig. \ref{fig_exvivo_dataset} shows selected sequential navigation results on both 2D and 3D end\footnote{Readers are strongly advised to watch the attached video to appreciate the performance.}.~\cite{behr2019deep}'s approach was re-implemented as an example of RL thanks to the manuscript's clear presentation. We followed~\cite{behr2019deep}'s description and enhanced it by using the last 4 frames as state and pretraining with the expert's operations. The rest setting strictly followed~\cite{behr2019deep}'s description. A qualitative comparison was also conducted to compare the performance of VascularPilot3D and a professional surgeon. The surgeon obtained an M.D. We denote one control loop as an ``imaging-localization-control-mechanic'' cycle (VascularPilot3D) and an ``imaging-control-mechanic'' cycle (human). \textbf{This work compares the number of control loops in each navigation trial.} It should be emphasized that time consumption cannot be fairly compared because of mechanical and imaging delay, and improvement in hardware will be our main focus in the future.\par
    
    Experiments show that VascularPilot3D achieves $100\%$ navigation success rate on all five episodes of each task. Table \ref{Table_exvivo_dataset} records the averages and standard deviations of 5 trial control loops in each task. It shows that VascularPilot3D outperforms the surgeon in 3 tasks out of 5 regarding control loops. RL navigation~\cite{behr2019deep} failed 5 out of 25 trials. Since \cite{behr2019deep}'s was implemented and trained by ourselves, and the setup was different, its result is for reference only. Based on the ex-vivo experiments, VascularPilot3D reduces $18.38\%$ surgeon's overall control loops. \par 
    
    One potential explanation for the better performance of VascularPilot3D is that when the guidewire enters the wrong bifurcation, VascularPilot3D identifies the error earlier than the human operator. In our recorded video, we observed that human operators notice and adjust the wrong forwarding to backwarding after 3-5 steps while VascularPilot3D moves backward immediately. Moreover, the complicated topology in the phantom misleads human operators, especially if he/she is not familiar with the phantom. This also happens in the tests RL method where the agent was stuck in the wrong turning points.\par

    The experiments also reveal that our ``trial-and-error'' and ``backward-and-forward'' guarantees the successful rate of the navigation procedure. Yet, we notice in some trials (task 1 and task 4) that it takes a long time for the guidewire to navigate through some turns. It should be credited to the issue of the intensities of translation and rotation motion controls being fixed. This may be solved by better coupling VascularPilot3D and hardware by fine-tuning these parameters.


    \subsection{Time consumption in loop control}
    
    VascularPilot3D consumes $550 ms$ total, and Otsu-based vessel segmentation is the bottleneck. Time consumption for vessel segmentation and 3D-2D alignment are around $500 ms$ and $50 ms$. Global and local path planning consume less than $45 ms$ and $10 ms$. Guidewire endpoint tracking costs $50 ms$ per frame and runs in parallel. In addition to modules in VascularPilot3D, the time consumptions in hardware were also summarized. Specifically, the mechanic takes around $1.3 - 1.8 s$ to fulfill the control message from VascularPilot3D. The imaging process takes around $200 ms$.

    \subsection{Limitations}
    \label{sec_limit}
    VascularPiolot3D can be further improved by four directions. First, an additional force feedback control can determine the range of translation value $\mathbf{c}$ for safety purposes and rapid parameter adaptation. Second, CPU-based Otsu-based segmentation (around $0.5 s$), mechanical (around $1.3 - 1.8 s$), and data transmission delay (around $0.2 s$) are the most time-consuming modules. By polishing our preliminary robot hardware and imaging procedure, the one-step control loop can be significantly improved from $2 s$ to less than $0.2 s$. Third, the collaboration of guidewire and catheter is reported to be essential for real-world clinical applications. Lastly, VascularPiolot3D can be further improved to optimize the direction of X-ray imaging to reduce radiation and leverage better 3D-2D registration. 
    

	\section{Conclusion}

    This work proposes the first 3D prior-free autonomous endovascular navigation system VascularPilot3D. VascularPilot3D is a complete system that infers the optimal sequential controls based on the pre-operative 3D and intra-operative continuous 2D fluoroscopy. Navigation in 3D space overcomes the ambiguous topological problem in existing 2D navigation works. VascularPilot3D has been coupled to our hardware system. Ex-vivo experiments on phantom validate that VascularPilot3D achieves $100\%$ success rate among 25 trials and reduces the surgeon's overall control loops by $18.38\%$. Moreover, VascularPilot3D is a parallel system to the interventional robots and imaging system. Considering that it is free of heavy hardware-related navigation training, which is essential in most RL systems, VascularPilot3D can be easily adapted and mounted on most EIGIs and imaging systems. \par
    Future works will be focused on adapting VascularPilot3D from ex-vivo to in-vivo experiments. The limitations listed in Section~\ref{sec_limit} should thus be addressed in further studies to ensure robust and safe navigation.\par

\balance
{\small
		\bibliographystyle{ieeetr}
		\bibliography{bib/strings-abrv,bib/ieee-abrv,annot}
	}

\end{document}